\documentclass{article} 
\usepackage{iclr2024_conference,times}

\usepackage{amsmath,amsfonts,bm}

\def\eqref#1{equation~\ref{#1}}

\def\1{\bm{1}}

\DeclareMathAlphabet{\mathsfit}{\encodingdefault}{\sfdefault}{m}{sl}
\SetMathAlphabet{\mathsfit}{bold}{\encodingdefault}{\sfdefault}{bx}{n}

\usepackage{hyperref}
\usepackage{url}
\usepackage{makecell}
\usepackage{booktabs}
\usepackage{amssymb}
\usepackage{graphicx}
\usepackage{amsmath,amssymb,amsfonts,bbding}
\usepackage{algorithmic}
\usepackage{textcomp}
\usepackage[caption=false,font=footnotesize]{subfig}
\usepackage{algorithm}
\usepackage{amsmath,bm}
\usepackage{multirow}
\usepackage{threeparttable}
\usepackage{color}
\usepackage{colortbl}
\usepackage{xcolor}
\usepackage{hyperref}
\usepackage{stfloats}
\usepackage{multirow}
\usepackage{adjustbox}
\usepackage{pifont}
\usepackage{comment}
\newcommand{\cmark}{\ding{51}}%
\newcommand{\xmark}{\ding{55}}%

\title{Efficient Remote Sensing with Harmonized Transfer Learning and Modality Alignment}

\author{Tengjun Huang\\
Shandong University\\
\texttt{htj@mail.sdu.edu.cn} \\
}

\iclrfinalcopy 
\begin{document}

\maketitle

\begin{abstract}
With the rise of Visual and Language Pretraining (VLP), an increasing number of downstream tasks are adopting the paradigm of pretraining followed by fine-tuning. Although this paradigm has demonstrated potential in various multimodal downstream tasks, its implementation in the remote sensing domain encounters some obstacles. Specifically, the tendency for same-modality embeddings to cluster together impedes efficient transfer learning. To tackle this issue, we review the aim of multimodal transfer learning for downstream tasks from a unified perspective, and rethink the optimization process based on three distinct objectives. We propose ``Harmonized Transfer Learning and Modality Alignment (HarMA)'', a method that simultaneously satisfies task constraints, modality alignment, and single-modality uniform alignment, while minimizing training overhead through parameter-efficient fine-tuning.  Remarkably, without the need for external data for training, HarMA achieves state-of-the-art performance in two popular multimodal retrieval tasks in the field of remote sensing. Our experiments reveal that HarMA achieves competitive and even superior performance to fully fine-tuned models with only minimal adjustable parameters. Due to its simplicity, HarMA can be integrated into almost all existing multimodal pretraining models. We hope this method can facilitate the efficient application of large models to a wide range of downstream tasks while significantly reducing the resource consumption\footnote{Code is available at \url{https://github.com/seekerhuang/HarMA}.}. 
\end{abstract}

\section{Introduction}\label{intro}
The advent of Visual and Language Pretraining (VLP) has spurred a surge in studies employing large-scale pretraining and subsequent fine-tuning for diverse multimodal tasks \citep{tan2019lxmert,li2020oscar,li2021align,li2022blip,liu2023visual}. When conducting transfer learning for downstream tasks in the multimodal domain, the common practice is to first perform large-scale pre-training and then fully fine-tune on a specific domain dataset \citep{hu2021unit,akbari2021vatt,zhang2024user}, which is also the case in the field of remote sensing image-text retrieval \citep{cheng2021deep,pan2023reducing}. However, this method has at least two notable limitations. Firstly, fully fine-tuning a large model is extremely expensive and not scalable \citep{zhang2024mmllms}. Secondly, the pre-trained model has already been trained on a large dataset for a long time, and fully fine-tuning on a small dataset may lead to reduced generalization ability or overfitting.

Recently, several works have attempted to use Parameter-Efficient Fine-Tuning (PEFT) to address this issue, aiming to freeze most of the model parameters and fine-tune only a few \citep{houlsby19a,mao2021unipelt,9760473}. This strategy seeks to incorporate domain-specific knowledge into the model while preserving the bulk of its original learned information. For example, \citet{houlsby19a} attempted to fine-tune the pre-trained model by simply introducing a single-modality MLP layer. In contrast, \citet{yuan2023parameter} designed a cross-modal interaction adapter, aiming to enhance the model's ability to integrate multimodal knowledge. Although the above works have achieved promising results, they either concentrate on single-modality features or overlook potential semantic mismatches when modeling the visual-language joint space.

\begin{figure}[htbp]
\centering
\includegraphics[width=0.6\textwidth]{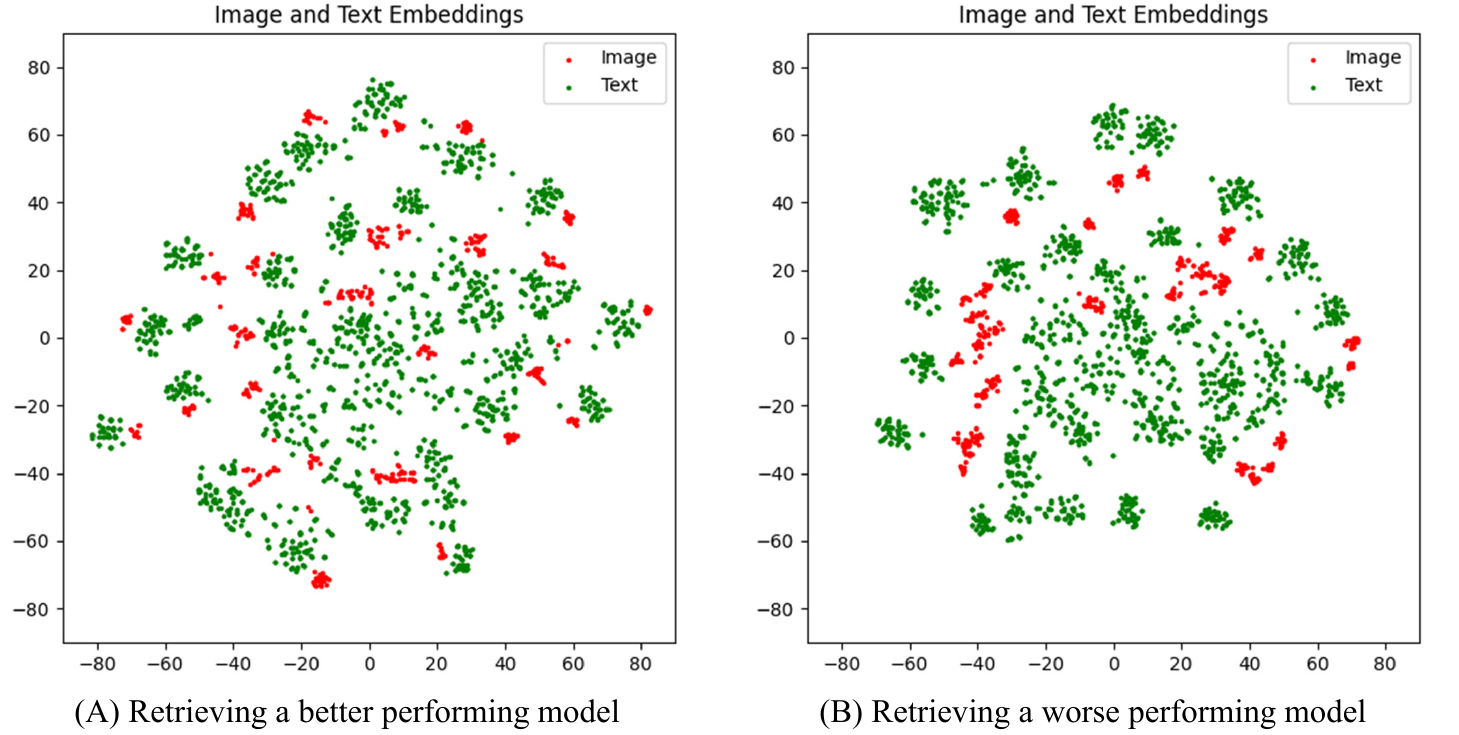}
\caption{In remote sensing image-text retrieval, excessive clustering of the same modality sometimes leads to a decrease in performance. The experiment was conducted on the RSITMD dataset.}
\label{TSNE}
\end{figure}

We have observed that poorly performing models sometimes exhibit a clustering phenomenon within the same modality embedding. Figure \ref{TSNE} illustrates the visualization of the last layer embeddings for two models with differing performance in the filed of remote sensing image-text retrieval; the clustering phenomenon is noticeably more pronounced in the right image than in the left. We hypothesize that this may be attributed to the high intra-class and inter-class similarity of remote sensing images, leading to semantic confusion when modeling a low-rank visual-language joint space. This raises a critical question: \textit{``How can we model a highly aligned visual-language joint space while ensuring efficient transfer learning?''}

In the brains of congenitally blind individuals, parts of the visual cortex can take on the function of language processing \citep{bedny2011language}. Concurrently, in the typical human cortex, several small regions—such as the Angular Gyrus and the Visual Word Form Area (VWFA)—serve as hubs for integrated visual-language processing \citep{10.1093/cercor/5.2.95}. These areas hierarchically manage both low-level and high-level stimuli information \citep{chen2019visual}. Inspired by this natural phenomenon, we propose ``Efficient Remote Sensing with \textbf{Har}monized Transfer Learning and \textbf{M}odality \textbf{A}lignment (HarMA)''. Specifically, similar to the information processing methods of the human brain, we designed a hierarchical multimodal adapter with mini-adapters. This framework emulates the human brain's strategy of utilizing shared mini-regions to process neural impulses originating from both visual and linguistic stimuli. It models the visual-language semantic space from low to high levels by hierarchically sharing multiple mini-adapters. Finally, we introduced a new objective function to alleviate the severe clustering of features within the same modality. Thanks to its simplicity, the method can be easily integrated into almost all existing multimodal frameworks.

\section{Method}

\subsection{Overall Framework}

\begin{figure}[htbp]
\centering
\includegraphics[width=1\textwidth]{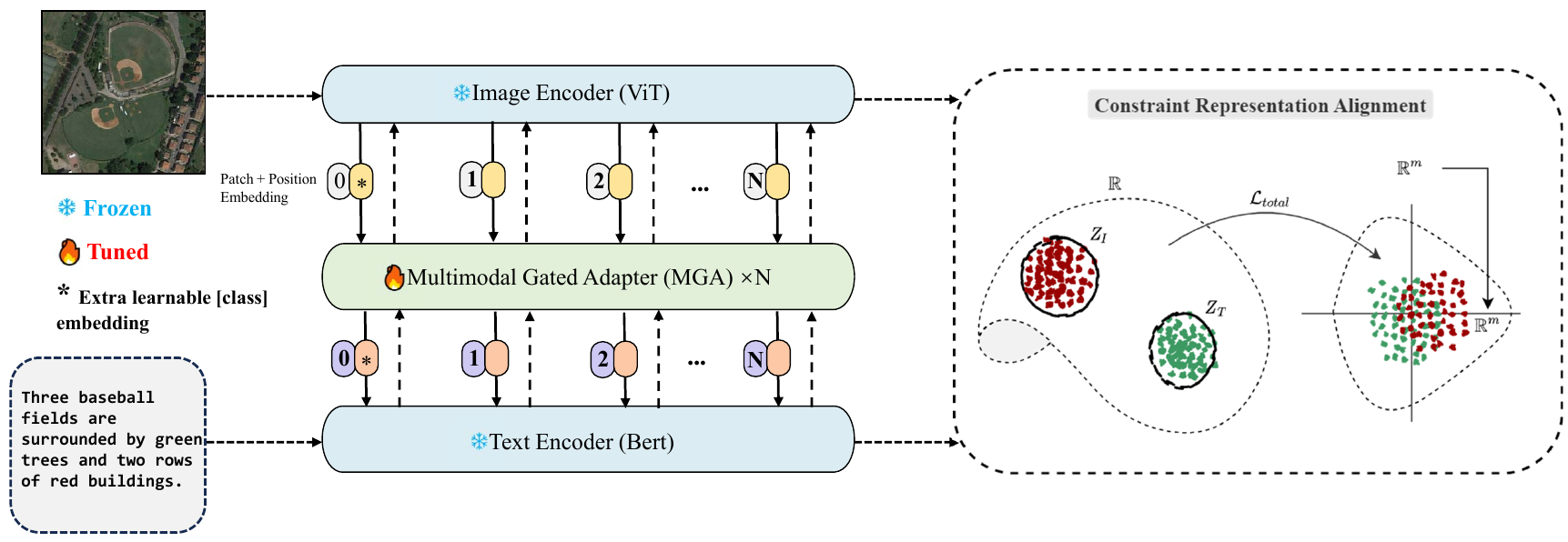}
\vspace{-24pt}
\caption{The overall framework of the proposed method.}
\label{str}
\vspace{-10pt}
\end{figure}
Figure \ref{str} illustrates our proposed HarMA framework. It initiates the process with the extraction of representations using image and text encoders, similar to CLIP \citep{radford2021learning}. These features are then processed by our unique multimodal gated adapter to obtain refined feature representations. Unlike the simple linear layer interaction used in \citep{yuan2023parameter}, we employ a shared mini adapter as our interaction layer within the entire adapter. After that, we optimize using a contrastive learning objective and our adaptive triplet loss.

\subsection{MultiModal Gated Adapter}

\begin{figure}[htbp]
\centering
\includegraphics[width=0.7\textwidth]{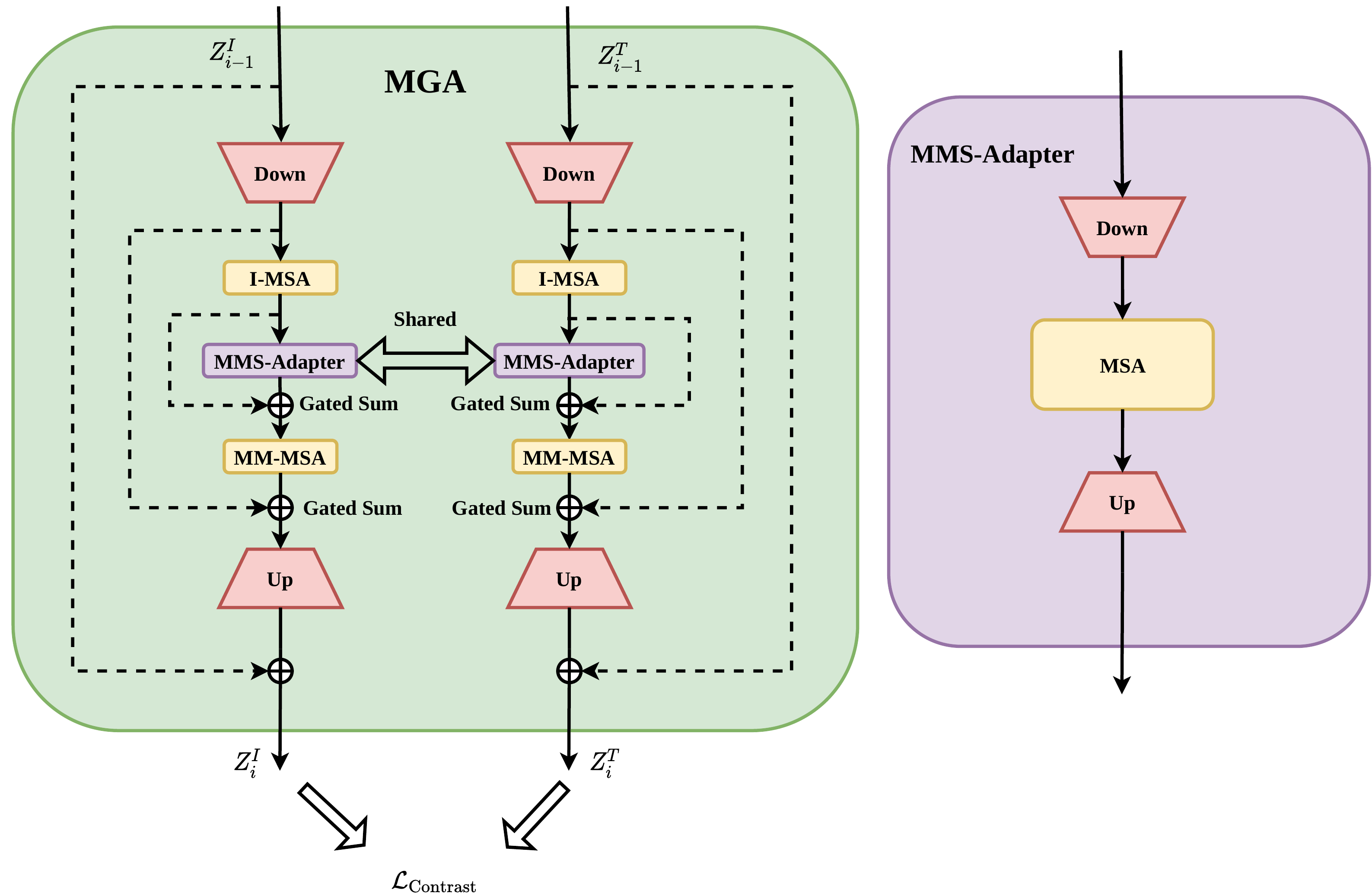}
\caption{The specific structure of the multimodal gated adapter.The overall structure is shown on the left, while the structure of the shared multimodal sub-adapter is displayed on the right.}
\label{module}
\end{figure}

Previous parameter-efficient fine-tuning methods in the multimodal domain \citep{jiang2022cross,yuan2023parameter} use a simple shared-weight method for modal interaction, potentially causing semantic matching confusion in the inherent modal embedding space. To address this, we designed a cross-modal adapter with an adaptive gating mechanism (Figure \ref{module}).

In this module, the extracted features $z_{I}$ and $z_{T}$ are first projected into low-dimensional embeddings. Different features $z_{I}$ ($z_{T}$) are further enhanced in feature expression after non-linear activation and subsequent processing by I-MSA. I-MSA and the subsequent MM-MSA share parameters. The features are then fed into our designed Multimodal Sub-Adapter (MMS-Adapter) for further interaction, the structure of this module is shown on the right side of Figure \ref{module}.

The MMS-Adapter, akin to standard adapters, aligns multimodal context representations via shared-weight self-attention. However, direct post-projection output of these aligned representations negatively impacts image-text retrieval performance, likely due to off-diagonal semantic key matches in the feature's low-dimensional manifold space. This contradicts contrastive learning objectives. 

To tackle this, the already aligned representations are further processed in the MSA with shared weights, thereby reducing model parameters and leveraging prior modality knowledge. And to ensure a finer-grained semantic match between image and text, we introduce early image-text matching supervision in the MGA output, significantly mitigating the occurrence of above issue.

Ultimately, features are projected back to their original dimensions before adding the skip connection. The final layer is initialized to zero to safeguard the performance of the pre-trained model during the initial stages of training. Algorithm \ref{alg:MGA} summarizes the proposed method.

\subsection{Objective Function}
In the realm of multimodal learning, when engaging in transfer learning for downstream tasks, it is typically necessary to devise objective functions tailored to distinct tasks and to align different modalities embeddings. we can initially define the objective of multimodal learning applied to all downstream tasks as follows:
\begin{equation}
    \min_{\theta^*}\left(\sum_i\mathbb{E}_{x_i\sim\mathcal{D}^i}[L_{\mathrm{task}}^i(f(x_i;\theta^*))]+\sum_{j\neq k}\mathbb{E}_{(x_j,x_k)\sim\mathcal{D}^j\times\mathcal{D}^k}[L_{\mathrm{align}}^{jk}(f(x_j;\theta^*),f(x_k;\theta^*))]\right).
    \label{obj1}
\end{equation}
Here, $L_{\mathrm{task}}^i$ represents the task loss for the $i$-th task, and $L_{\mathrm{align}}^{jk}$ denotes the alignment loss between different pairs of modalities $(j, k)$. The expectation is taken over the data distribution $\mathcal{D}$ for each task. $\theta^*$  represents the target parameters for transfer learning.

However, in the remote sensing field, we observe that underperforming models sometimes exhibit a phenomenon where same-modality embeddings cluster together, as shown in Figure \ref{TSNE}. \citet{wang2020understanding} highlighted that the low uniformity of alignment of modalities from the same distribution may limit the transferability of embeddings. To ensure that embeddings from the same modality are uniformly aligned without excessive clustering, the unified objective for multimodal learning applied to remote sensing downstream tasks can be defined as:
\begin{equation}
    \begin{gathered}
\min_{\theta^*}\left(L_{\mathrm{ini}}+\lambda_1\sum_i\mathbb{E}_{x_i\sim\mathcal{D}^i}[L_{\mathrm{uniform}}^i(f(x_i;\theta^*))]\right) \\
\mathrm{s.t.~}D(\theta,\theta^{*})\leq\delta  .
\end{gathered}
\end{equation}
In this equation, $L_{\mathrm{ini}}$ represents the initial optimization objective (Equation \ref{obj1}), which is composed of the task loss and alignment loss. $L_{\mathrm{uniform}}^i$ denotes the singe-modality uniformity loss for the $i$-th modality, and $D(\theta,\theta^{*})$ is a cost measure between the original and updated model parameters, constrained to be less than $\delta$.  $\delta$ is the minimum parameter update cost in the ideal state. 

We observe that existing works often only explore one or two objectives, with most focusing either on how to efficiently fine-tune parameters for downstream tasks \citep{jiang2022cross,jie2022convolutional,yuan2023parameter} or on modality alignment \citep{chen2020uniter,ma2023direction,pan2023prior}. Few can simultaneously satisfy the three requirements outlined in the above formula. We have satisfied the need for efficient transfer learning by introducing adapters that mimic the human brain. This prompts us to ask: how can we fulfill the latter two objectives—high alignment of embeddings across different modalities while preventing excessive clustering of embeddings within the same modality?

\subsubsection{Adaptive Triplet Loss}
In image-text retrieval tasks, the bidirectional triplet loss established by \citet{faghri2017vse++} has become a mainstream loss function. However, if we choose to accumulate the losses of all samples in image-text matching, the model may struggle to optimize due to the high intra-class similarity and inter-class similarity prevalent in most regions of the images within the field of remote sensing \citep{yuan2022exploring}. Therefore, We propose an Adaptive Triplet Loss that automatically mines and optimizes hard samples:
\begin{equation}
    \mathcal{L}_{\text{ada-triplet}} = \frac{1}{2}\left(\sum_{i=1}^{N}\sum_{j=1}^{N} w_{s, ij} \left[m + s_{ij} - s_{ii}\right]_+ + \sum_{j=1}^{N}\sum_{i=1}^{N} w_{s, ji} \left[m + s_{ji} - s_{ii}\right]_+\right) ,
\end{equation}

where $s_{ij}$ is the dot product between image feature $i$ and text feature $j$, $w_i$ and $w_j$ are the weights of sample $i$ and $j$, determined by the loss size of different samples:

\begin{equation}
    w_{s, ij} = (1 - \exp(-[m + s_{ij} - s_{ii}]_+))^{\gamma},w_{s, ji} = (1 - \exp(-[m + s_{ji} - s_{ii}]_+))^{\gamma},
\end{equation}

where $\gamma$ is a hyperparameter adjusting the size of the weights. This loss function aims to bring the features of positive samples closer together, while distancing those between positive and negative samples. By dynamically adjusting the focus between hard and easy samples, our approach effectively satisfies the other two objectives proposed above. It not only aligns different modality samples at a fine-grained level but also prevents over-aggregation among samples of the same modality, thereby enhancing the model's matching capability. Also, following the approach of \citep{radford2021learning}, we utilize a contrastive learning objective to align image and text semantic features. Consequently, the total objective is defined as:
\begin{equation}
    \mathcal{L}_{\text{total}} = (\lambda_{1}\mathcal{L}_{\text{ada-triplet}} +\lambda_{2}\mathcal{L}_{\text{contrastive}}) .
\end{equation}
The $\lambda_{1}$ and $\lambda_{2}$ are parameters for balancing the loss. We offer detailed information on contrastive learning in Appendix \ref{cl}.

\section{Experiments}
We evaluate our proposed HarMA framework on two widely used remote sensing (RS) image-text datasets: RSICD \citep{lu2017exploring} and RSITMD \citep{yuan2022exploring}. We use standard recall at TOP-K (R@K, K = 1, 5, 10) and mean recall (mR) to assess our model.

\subsection{Comparative Experiments}

In this section, we compare the proposed method with state-of-the-art retrieval techniques on two remote sensing multimodal retrieval benchmarks. The backbone networks employed in our experiments are the CLIP (ViT-B-32) \citep{pmlr-v139-radford21a} and the GeoRSCLIP \citep{zhang2023rs5m}.

\vspace{-10pt}
\begin{table}[htbp]
\caption{Retrieval Performance Test. 
$\ddagger$ : RSICD Test Set; 
$\divideontimes$ : RSITMD Test Set; $\dag$ : The parameter count of a single adapter module, and $\dag'$ represents the results referenced from \citet{yuan2023parameter}. \textcolor{red}{Red}: Our method; \textcolor{blue}{Blue}: Full fine-tuned CLIP \& GeoRSCLIP.}
\vspace{5pt}
\footnotesize
\setlength{\tabcolsep}{3pt} 
\begin{adjustbox}{width=\textwidth,center}
\begin{tabular}{lccccccccc}
\hline
     &   &    & \multicolumn{3}{c}{\textbf{Image-to-text}}                                                                       & \multicolumn{3}{c}{\textbf{Text-to-image}}                                                                      &                               \\
\multirow{-2}{*}{\textbf{Methods}}  &  \multirow{-2}{*}{\textbf{Backbone (image/text)}} &  \multirow{-2}{*}{\textbf{Trainable Params}}   & \textbf{R@1}                  & \textbf{R@5}                  & \textbf{R@10}                 & \textbf{R@1}                  & \textbf{R@5}                  & \textbf{R@10}                 & \multirow{-2}{*}{\textbf{mR}} \\ \hline
\multicolumn{10}{l}{\textbf{\textit{\textbf{Traditional methods}}}} \\ \hline
GaLR with MR \citep{9745546}      & ResNet18, biGRU $\ddagger$ & 46.89M         & 6.59                 & 19.85                & 31.04                & 4.69                          & 19.48                         & 32.13                         & 18.96                \\ 
PIR \citep{pan2023prior}      & Swin Transformer, Bert $\ddagger$  & -         & 9.88                 & 27.26                & 39.16                & 6.97                          & 24.56                         & 38.92                        & 24.46                \\ \hline
\multicolumn{10}{l}{\textbf{\textit{\textbf{CLIP-based methods}}}} \\ \hline

Full-FT CLIP \citep{pmlr-v139-radford21a}    &  CLIP(ViT-B-32) $\ddagger$      & 151M    & \textcolor{blue}{\textbf{15.89}}                & \textcolor{blue}{\textbf{36.14}}                & \textcolor{blue}{\textbf{47.93}}                & \textcolor{blue}{\textbf{12.21}}                & \textcolor{blue}{\textbf{32.97}}                & \textcolor{blue}{\textbf{48.84}}                & \textcolor{blue}{\textbf{32.33}}                \\ 

Full-FT GeoRSCLIP \citep{zhang2023rs5m}    &  GeoRSCLIP(ViT-B-32-RET-2) $\ddagger$      & 151M    & \textcolor{blue}{\textbf{18.85}}                & \textcolor{blue}{\textbf{38.15}}                & \textcolor{blue}{\textbf{53.16}}                & \textcolor{blue}{\textbf{14.27}}                & \textcolor{blue}{\textbf{39.71}}                & \textcolor{blue}{\textbf{57.49}}                & \textcolor{blue}{\textbf{36.94}}                \\ 
Full-FT GeoRSCLIP (w/ Extra Data)     &  GeoRSCLIP(ViT-B-32-RET-2) $\ddagger$      & 151M    & \textcolor{blue}{\textbf{21.13}}                & \textcolor{blue}{\textbf{41.72}}                & \textcolor{blue}{\textbf{55.63}}                & \textcolor{blue}{\textbf{15.59}}                & \textcolor{blue}{\textbf{41.19}}                & \textcolor{blue}{\textbf{57.99}}                & \textcolor{blue}{\textbf{38.87}}                \\ \hline
Adapter \citep{houlsby19a}           &  CLIP(ViT-B-32) $\ddagger$  &  0.17M $\dag'$          & 8.73                              & 24.73                              & 37.81                              & 8.43                              & 26.02                              & 43.33                             & 24.84                              \\
CLIP-Adapter \citep{10.48550}    &  CLIP(ViT-B-32) $\ddagger$  & 0.52M $\dag'$           & 7.11                          & 19.48                         & 31.01                         & 7.67                          & 24.87                         & 39.73                         & 21.65                         \\
AdaptFormer \citep{chen2022adaptformer}      &   CLIP(ViT-B-32) $\ddagger$ &  0.17M $\dag'$         & 12.46                              & 28.49                              & 41.86                              & 9.09                              & 29.89                              & 46.81                              & 28.10                             \\  
 Cross-Modal Adapter \citep{Jiang10.48550}      &   CLIP(ViT-B-32) $\ddagger$ & 0.16M $\dag'$            & 11.18                             & 27.31                              & 40.62                              & 9.57                              & 30.74                             & 48.36                              & 27.96                             \\
UniAdapter \citep{2302.06605}       &  CLIP(ViT-B-32) $\ddagger$  & 0.55M $\dag'$          & 12.65                             & 30.81                              & 42.74                              & 9.61                              & 30.06                             & 47.16                              & 28.84                             \\  
PE-RSITR \citep{yuan2023parameter}       &  CLIP(ViT-B-32) $\ddagger$  & 0.16M $\dag'$          & 14.13                             & 31.51                              & 44.78                              & 11.63                              & 33.92                             & 50.73                              & 31.12                             \\
\rowcolor{gray!20} Ours (HarMA w/o Extra Data)  & CLIP(ViT-B-32) $\ddagger$ & 0.50M $\dag$  & \textcolor{red}{\textbf{16.36}} & \textcolor{red}{\textbf{34.48}} & \textcolor{red}{\textbf{47.74}} & \textcolor{red}{\textbf{12.92}} & \textcolor{red}{\textbf{37.17}} & \textcolor{red}{\textbf{53.07}} & \textcolor{red}{\textbf{33.62}} \\
\rowcolor{gray!20} Ours (HarMA w/o Extra Data)  & GeoRSCLIP(ViT-B-32-RET-2) $\ddagger$ & 0.50M $\dag$  & \textcolor{red}{\textbf{20.52}} & \textcolor{red}{\textbf{41.37}} & \textcolor{red}{\textbf{54.66}} & \textcolor{red}{\textbf{15.84}} & \textcolor{red}{\textbf{41.92}} & \textcolor{red}{\textbf{59.39}} & \textcolor{red}{\textbf{38.95}} \\ \hline

Full-FT CLIP  \citep{pmlr-v139-radford21a}    &  CLIP(ViT-B-32) $\divideontimes$       & 151M    & \textcolor{blue}{\textbf{26.99}}                & \textcolor{blue}{\textbf{46.9}}                & \textcolor{blue}{\textbf{58.85}}                & \textcolor{blue}{\textbf{20.53}}                & \textcolor{blue}{\textbf{52.35}}                & \textcolor{blue}{\textbf{71.15}}                & \textcolor{blue}{\textbf{46.13}}                \\ 
Full-FT GeoRSCLIP \citep{zhang2023rs5m}    &  GeoRSCLIP(ViT-B-32-RET-2) $\divideontimes$       & 151M    & \textcolor{blue}{\textbf{30.53}}                & \textcolor{blue}{\textbf{49.78}}                & \textcolor{blue}{\textbf{63.05}}                & \textcolor{blue}{\textbf{24.91}}                & \textcolor{blue}{\textbf{57.21}}                & \textcolor{blue}{\textbf{75.35}}                & \textcolor{blue}{\textbf{50.14}}               \\
Full-FT GeoRSCLIP (w/ Extra Data)     &  GeoRSCLIP(ViT-B-32-RET-2) $\divideontimes$       & 151M    & \textcolor{blue}{\textbf{32.30}}                & \textcolor{blue}{\textbf{53.32}}                & \textcolor{blue}{\textbf{67.92}}                & \textcolor{blue}{\textbf{25.04}}                & \textcolor{blue}{\textbf{57.88}}                & \textcolor{blue}{\textbf{74.38}}                & \textcolor{blue}{\textbf{51.81}}                \\ \hline

CLIP-Adapter \citep{10.48550}    &  CLIP(ViT-B-32) $\divideontimes$  & 0.52M $\dag'$            & 12.83                          & 28.84                         & 39.05                         & 13.30                          & 40.20                        & 60.06                         & 32.38                         \\

AdaptFormer \citep{chen2022adaptformer}      &   CLIP(ViT-B-32) $\divideontimes$ &  0.17M $\dag'$         & 16.71                              & 30.16                              & 42.91                              & 14.27                              & 41.53                              & 61.46                             & 34.81                            \\  

Cross-Modal Adapter \citep{Jiang10.48550}      &   CLIP(ViT-B-32) $\divideontimes$ & 0.16M $\dag'$            & 18.16                             & 36.08                             & 48.72                              & 16.31                              & 44.33                             & 64.75                              & 38.06                             \\

UniAdapter \citep{2302.06605}       &  CLIP(ViT-B-32) $\divideontimes$   & 0.55M $\dag'$          & 19.86                             & 36.32                              & 51.28                              & 17.54                              & 44.89                             & 56.46                              & 39.23                             \\

PE-RSITR \citep{yuan2023parameter}       &  CLIP(ViT-B-32) $\divideontimes$   & 0.16M $\dag'$          & 23.67                             & 44.07                              & 60.36                              & 20.10                              & 50.63                             & 67.97                              & 44.47                             \\

\rowcolor{gray!20} Ours (HarMA w/o Extra Data)  & CLIP(ViT-B-32) $\divideontimes$  & 0.50M $\dag$  & \textcolor{red}{\textbf{25.81}} & \textcolor{red}{\textbf{48.37}} & \textcolor{red}{\textbf{60.61}} & \textcolor{red}{\textbf{19.92}} & \textcolor{red}{\textbf{53.27}} & \textcolor{red}{\textbf{71.21}} & \textcolor{red}{\textbf{46.53}} \\

\rowcolor{gray!20} Ours (HarMA w/o Extra Data)  & GeoRSCLIP(ViT-B-32-RET-2) $\divideontimes$  & 0.50M $\dag$  & \textcolor{red}{\textbf{32.74}} & \textcolor{red}{\textbf{53.76}} & \textcolor{red}{\textbf{69.25}} & \textcolor{red}{\textbf{25.62}} & \textcolor{red}{\textbf{57.65}} & \textcolor{red}{\textbf{74.60}} & \textcolor{red}{\textbf{52.27}} \\ \hline
\end{tabular}
\end{adjustbox}
\label{tab:results}
\end{table}
Table \ref{tab:results} presents the retrieval performance on RSICD and RSITMD. Firstly, as indicated in the first column, our method surpasses traditional state-of-the-art approaches while requiring significantly fewer tuned parameters. Secondly, when using CLIP (ViT-B-32) \citep{pmlr-v139-radford21a} as the backbone, our approach achieves competitive and even superior performance compared to fully fine-tuned methods. Specifically, when matched with methods that have a similar number of tunable parameters, our method's Mean Recall (MR) sees an approximate increase of 50\% over CLIP-Adapter \citep{10.48550} and 12.7\% over UniAdapter \citep{2302.06605} on RSICD, and an 18.6\% improvement over UniAdapter on RSITMD.
Remarkably, by utilizing the pretrained weights of GeoRSCLIP, HarMA establishes a new benchmark in the remote sensing field for two popular multimodal retrieval tasks. It only modifies less than 4\% of the total model parameters, outperforming all current parameter-efficient fine-tuning methods and even surpassing the image-text retrieval performance of fully fine-tuned GeoRSCLIP on RSICD and RSITMD.

\section{Conclusion}
In this paper, we have revisited the learning objectives of multimodal downstream tasks from a unified perspective and proposed HarMA, an efficient framework that addresses the suboptimal multimodal alignment in remote sensing. HarMA uniquely enhances uniform alignment while preserving pretrained knowledge. Through the use of lightweight adapters and adaptive losses, HarMA achieves state-of-the-art retrieval performance with minimal parameter updates, surpassing even full fine-tuning. Despite all the benefits, one potential limitation is that designing pairwise objective functions may not provide more robust distribution constraints. Our future work will focus on extending this approach to more multimodal tasks.

\bibliography{iclr2024_conference}
\bibliographystyle{iclr2024_conference}

\newpage
\appendix
\section*{Appendix}

\section{Preliminaries}\label{preli}

\subsection{Vision Transformer (ViT)}
In the pipeline of ViT \citep{dosovitskiy2020image}, given an input image \( I \) in \( \mathbb{R}^{C \times H \times W} \) segmented into \( N \times N \) patches, we define the input matrix \( \mathbf{X} \in \mathbb{R}^{(N^2 + 1) \times D} \) consisting of patch embeddings and a class token. The self-attention mechanism in each Transformer layer transforms $\mathbf{X}$ into keys \( K \), values \( V \), and queries \( Q \), each in \( \mathbb{R}^{(N^2 + 1) \times D} \), and computes the self-attention \citep{vaswani2017attention} output as:
\begin{equation}
\text{Attention}(Q, K, V) = \text{Softmax}\left(\frac{QK^T}{\sqrt{D}}\right)V.
\end{equation}
The subsequent processing involves a two-layer MLP, which refines the output across the embedded dimensions. 

\subsection{Adapter}
Adapters \citep{houlsby19a} are modular additions to pre-trained models, designed to allow adaptation to new tasks by modifying only a small portion of the model's parameters. An adapter consists of a sequence of linear transformations and a non-linear activation:
\begin{equation}
    \text{Adapter}(\mathbf{Z}) = [\mathbf{W}^\mathrm{up}\sigma(\mathbf{W}^\mathrm{down}\mathbf{Z}^T)]^T ,
\end{equation}
where \(\mathbf{W}^{\text{down}}\) and \(\mathbf{W}^{\text{up}}\) are the weights for down-projection and up-projection, while $\sigma$ denotes a non-linear activation function that introduces non-linearity into the model.

Integrating adapters into the ViT architecture allows for efficient fine-tuning on specific tasks while keeping the majority of the Transformer's weights unchanged. 
This approach not only retains the pre-trained visual understanding but also reduces the risk of catastrophic forgetting.
\subsection{Triplet Loss}

In the domain of image-text retrieval tasks, the bidirectional triplet loss introduced by \citet{faghri2017vse++} has been widely adopted as a standard loss function. The formulation of the triplet loss is given by:

\begin{equation}
    \mathcal{L}_{\text{triplet}} = \frac{1}{2}\left(\sum_{i=1}^{N}\sum_{j=1}^{N} \left[m + s_{ij} - s_{ii}\right]_+ + \sum_{j=1}^{N}\sum_{i=1}^{N} \left[m + s_{ji} - s_{ii}\right]_+\right) .
\end{equation}

The $m$ denotes the margin, a hyperparameter that defines the minimum distance between the non-matching pairs. The function $s_{ij}$ represents the similarity score between the $i$-th image and the $j$-th text, and $s_{ii}$ is the similarity score between matching image-text pairs. The objective of the triplet loss is to ensure that the distance between non-matching pairs is greater than the distance between matching pairs by at least the margin $m$.

\subsection{Contrastive Learning}\label{cl}


Given a mini-batch of $N$ positive image-text pairs $\{(I_i,T_i)\}_{i\in\{1,...,N\}}$, we obtain the final image and text embeddings $\{(v_i,t_i)\}_{i\in\{1,...,N\}}$, and compute the cosine similarity $\text{sim}(v_i, t_j) =v_i^\mathrm{T}t_j$. The vision-to-text and text-to-vision contrastive losses are defined as:

\begin{equation}
\begin{aligned}
    \mathcal{L}_{v2t} &= -\frac{1}{N} \sum_{i=1}^N \log \frac{\exp(\text{sim}(v_i, t_i) / \tau)}{\sum_{j=1}^N \exp(\text{sim}(v_i, t_j) / \tau)}, \\
    \mathcal{L}_{t2v} &= -\frac{1}{N} \sum_{i=1}^N \log \frac{\exp(\text{sim}(t_i, v_i) / \tau)}{\sum_{j=1}^N \exp(\text{sim}(t_i, v_j) / \tau)},
\end{aligned}
\end{equation}

where $\tau$ is the temperature parameter that scales the distribution of similarities. The overall contrastive loss is computed as the average of the vision-to-text and text-to-vision losses:

\begin{equation}
\mathcal{L}_{\text{contrastive}}=\frac{1}{2}\sum_{i=1}^{N}\left(\mathcal{L}_{v2t}+\mathcal{L}_{t2v}\right).
\end{equation}

This loss function encourages the model to distinguish between positive pairs (where images and texts correspond to each other) and negative pairs (where they do not), effectively learning a joint embedding space where similar concepts are closer together.

\subsection{Retrieval Evaluation Metrics}\label{metrics}

The retrieval evaluation metrics are defined as the mean of the recall rates at different cutoff points for both text-to-image and image-to-text retrieval tasks:

\begin{equation}
    \mathrm{mR}=\left(\underbrace{R@1+R@5+R@10}_{\text{Text-to-image}}+\underbrace{R@1+R@5+R@10}_{\text{Image-to-text}}\right)/6,
\end{equation}

where $R@k$ is the recall at rank $k$, indicating the percentage of queries for which the correct item is found among the top $k$ retrieved results. The mean recall $\mathrm{mR}$ averages these recall values to provide a single performance metric that captures retrieval effectiveness at various depths of the result list.

\section{Algorithm}\label{algo}

\begin{algorithm}[htbp]
\caption{\label{alg:MGA} MultiModal Gated Adapter (MGA) for cross-modal interaction.}
\begin{algorithmic}
    \STATE \textbf{Input:} Feature tensors $Z_I$ and $Z_T$ from image and text encoders, respectively.
    \STATE \textbf{Parameters:} Weight matrices $\mathbf{W_1}, \mathbf{W_2}, \mathbf{W_i}$, bias vectors $\mathbf{b_1}, \mathbf{b_2}, \mathbf{b_i}$, and learnable gate parameters $\lambda_1, \lambda_2$.
    \STATE \textbf{Output:} Enhanced feature tensors $f_{end}^I$ and $f_{end}^T$ for image and text.
    
    \STATE \textbf{function} $\sigma(\cdot)$ \textbf{is} non-linear activation function (e.g., GELU)
    \STATE \textbf{function} $MSA(\cdot)$ \textbf{is} Multi-Head Self-Attention mechanism
    \STATE \textbf{function} $MMSA(x)$ \textbf{is} Multi-Modal Sub-Adapter mechanism defined as:
    \STATE \quad $MMSA(x) =  \mathbf{W_i}^\mathbf{Up}(MSA(\sigma(\mathbf{W_i}^\mathbf{Down}x + \mathbf{b_i}^\mathbf{Down}))) + \mathbf{{b_i}^{Up}}$
    
    \FOR{each feature tensor $Z$ in $\{Z_I, Z_T\}$}
        \STATE $f_1 = \sigma(\mathbf{W_1}Z + \mathbf{b_1})$ \textcolor{gray}{\# Process image and text feature tensors}
        \STATE $f_2 = MSA(f_1)$
        \STATE $f_3 = \lambda_1 MMSA(f_2) + (1 - \lambda_1)f_2$ \textcolor{gray}{\# Apply multi-modal sub-adapter with gating}
        \STATE $f_4 = \lambda_2 MSA(f_3) + (1 - \lambda_2)f_1$
        \STATE $f_{end} = (\mathbf{W_2}f_4 + \mathbf{b_2}) + Z$
    \ENDFOR
    
    \STATE \textbf{return} $f_{end}^I$, $f_{end}^T$
\end{algorithmic}
\end{algorithm}

\section{Experiment Details}\label{expdetails}
\subsection{Datasets}
\textbf{RSICD.} The Remote Sensing Image Captioning Dataset (RSICD) serves as a benchmark for the task of captioning remote sensing images \citep{lu2017exploring}. It encompasses over ten thousand remote sensing images sourced from Google Earth, Baidu Maps, MapABC, and Tianditu. The resolution of these images varies, with each being resized to a fixed dimension of 224x224 pixels. The dataset comprises a total of 10,921 images, with each image accompanied by five descriptive sentences.

\textbf{RSITMD.} The Remote Sensing Image-Text Matching Dataset (RSITMD) is a fine-grained and challenging dataset for remote sensing multimodal retrieval tasks, introduced by \citet{yuan2022exploring}. Unlike other remote sensing image-text pairing datasets, it features detailed descriptions of the relationships between objects. Additionally, the dataset includes keyword attributes (ranging from one to five keywords per image), facilitating keyword-based remote sensing text retrieval tasks. It comprises 4,743 images spanning 32 scenes, with a total of 23,715 annotations, of which 21,829 are unique.

\subsection{Implementation Details}
Consistent with the experimental details established by \citet{pan2023prior}, we partitioned the dataset into distinct sets for training, validation, and testing. For CLIP (ViT-B-32)\footnote{\url{https://github.com/mlfoundations/open_clip}.} or GeoRSCLIP, the final features are linearly projected to a 512-dimensional space. We set the temperature coefficient for the contrastive loss at 0.07 and the margin for the adaptive triplet loss at 0.2. Training was executed on either four A40 GPUs (48GB × 2) or eight RTX 4090 GPUs (24GB × 4) with a batch size of 428. We utilized an AdamW optimizer \citep{loshchilov2017decoupled} with a learning rate of 4e-4 (4e-6 for fine-tuning GeoRSCLIP), a weight decay of 0.04, and implemented a linear decay strategy for the learning rate. Following the approach of \citet{yuan2022exploring}, we employ a k-fold cross-validation strategy to obtain average results, with \( k \) set to 5.
\subsection{Ablation Study}

To demonstrate the effectiveness of our proposed method, we compared it against three baselines: 1) a CLIP model fully fine-tuned with downstream domain data, 2) a CLIP model with MGA but without introducing a new objective function, and 3) a CLIP model with HarMA. We evaluate the experimental performance on the RSITMD dataset. Table \ref{tab:method_comparison} presents the results of the ablation study.
\begin{table}[ht]
\vspace{-15pt}
\centering
\caption{Comparison of different methods on RSITMD dataset. The first row represents the results of the fully fine-tuned CLIP (Full-ft CLIP). The subsequent rows, from top to bottom, represent the CLIP with the Multimodal Gating Adapter (MGA) where only the MGA is fine-tuned (CLIP w/ MGA), and CLIP with both the MGA and the adaptive triplet loss, where only the MGA is fine-tuned (CLIP w/ MGA + Adaptive Triplet Loss).}
\label{tab:method_comparison}
\scriptsize 
\begin{tabular}{l|c|cc|ccc|ccc|c}
  \toprule 
  \multirow{2}{*}{Method} & \multirow{2}{*}{Backbone} & \multicolumn{2}{c|}{Module} & \multicolumn{3}{c|}{Text-to-image} & \multicolumn{3}{c|}{Image-to-text} & \multirow{2}{*}{\centering MR  } \\
  & & MGA & Adaptive Triplet Loss & R@1 & R@5 & R@10 & R@1 & R@5 & R@10 & \\ \hline
  Full-FT CLIP & ViT-B-32 & \xmark & \xmark & 26.99 & 46.9 & 58.85 & 20.53 & 52.35 & 71.15 & \multicolumn{1}{c}{46.13}  \\ \hline
  CLIP w/ MGA  & ViT-B-32 & \cmark & \xmark & 25.33 & 47.96 & 60.26 & 18.71 & 53.13 & 70.52 & \multicolumn{1}{c}{45.98}  \\
  CLIP (HarMA) & ViT-B-32 & \cmark & \cmark & 25.81 & 48.37 & 60.61 & 19.92 & 53.27 & 71.21 & \multicolumn{1}{c}{46.53}  \\
  \bottomrule
\end{tabular}
\end{table}

\begin{table}[htbp]
\vspace{-20pt}
\centering
\caption{Model trainable parameters comparison.}
\label{tab:performance_comparison}
\begin{tabular}{ccc}
\toprule
 Method & Trainable Params (\%)  \\
\midrule
Full-FT CLIP & 100.00  \\
Adapter \citep{houlsby19a} & 2.60  \\
CLIP-Adapter \citep{10.48550} & 4.13  \\
 CLIP w/ MGA (Ours) & 3.82\\
 HarMA (Ours)   & 3.82      \\
\bottomrule
\end{tabular}
\end{table}

In our experiments, for a fair comparison, we selected CLIP (ViT-B-32) \citep{radford2021learning} as the backbone model to conduct ablation studies. As shown in Table \ref{tab:method_comparison}, the first row presents the performance of CLIP following comprehensive fine-tuning on downstream datasets, achieving a robust performance (mR=46.13). However, this method requires fine-tuning the entire model, which is computationally intensive and difficult to scale. By simply integrating the Multimodal Gated Adapter (MGA) as depicted in the third row, we achieve results comparable to full fine-tuning (45.98 vs 46.13). The third row illustrates that the inclusion of our proposed adaptive triplet loss further enhances performance, slightly surpassing that of full fine-tuning (46.53 vs 46.13). This improvement can be attributed to the introduction of the objective function that addresses the issue of semantic confusion within the same modality, as discussed in Section \ref{intro}, when modeling the joint visual-language space. Our experiments validate the effectiveness of the proposed modules, offering a promising solution to address this challenge. Table \ref{tab:performance_comparison} presents the percentage of parameters that require fine-tuning for each method, relative to the total number of parameters in the model.

\section{Qualitative Analysis}

\subsection{Image-to-Text Results Analysis}

\begin{figure}[htbp]
\centering
\includegraphics[width=1.0\textwidth]{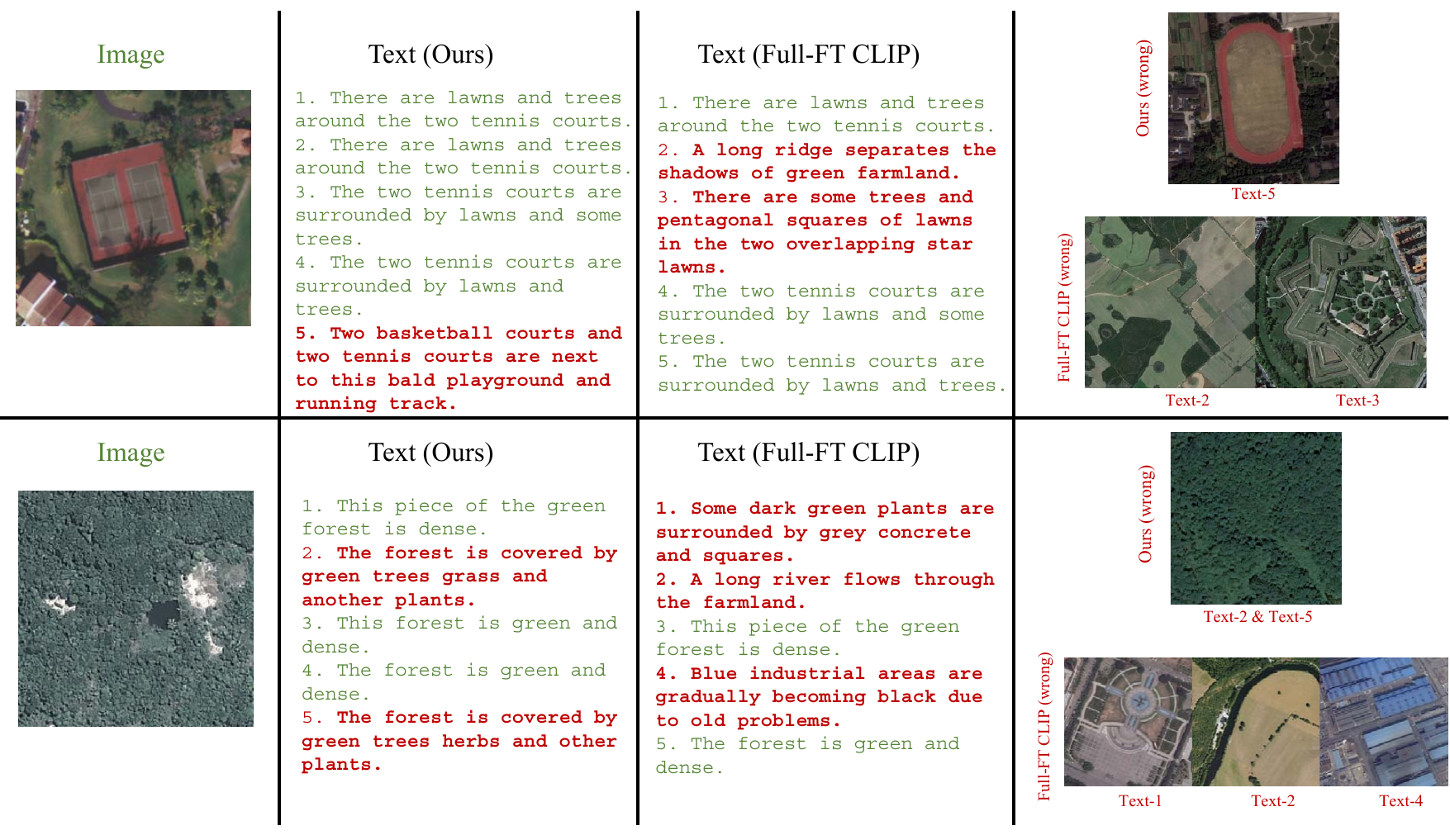}
\caption{\textbf{Image-to-Text Retrieval Visual Results.} We compare the top-5 retrieved captions from HarMA (Ours) and the fully fine-tuned CLIP (Full-FT CLIP). Green text indicates correct retrieval, while red text indicates incorrect retrieval. The image on the far right corresponds to the text where retrieval errors occurred. Overall, HarMA demonstrates superior retrieval performance compared to Full-FT CLIP. Our method accurately captures the semantic elements, such as the tennis court in the first image, and associates them with the overall context. In contrast, Full-FT CLIP tends to overemphasize irrelevant details, mistakenly treating partial shadows and surrounding trees as the main subject matter.}
\label{i2t res}
\end{figure}

Figure \ref{i2t res} presents the qualitative results for image-to-text retrieval, comparing the top-5 retrieved captions from HarMA (Ours) and the fully fine-tuned CLIP (Full-ft CLIP). Overall, HarMA demonstrates superior retrieval performance compared to Full-ft CLIP. 

Notably, in the first case, even though our method's fifth-ranked retrieval result is incorrect, it effectively captures the overall semantics rather than irrelevant details. For instance, we identify the ``two tennis courts'' as the core semantic element of the image and retrieve a highly similar text (see the incorrect example in the rightmost column). In contrast, Full-ft CLIP mistakenly treats the ``slight shadows'' of the tennis courts and the surrounding ``trees and lawn'' as the primary semantics, as evidenced by the second (shadow of green farmland) and third (some trees and pentagonal squares of lawns) retrieved texts. This observation may provide two insights from an interpretable perspective: \textbf{1) HarMA is better at recognizing overall semantics, and 2) Full-ft CLIP may be prone to overfitting, focusing excessively on unimportant details in the image.}

In the second scenario, we focus on a forest image, which poses a greater challenge. Unlike the first scenario, here trees typically seen in the background are brought to the foreground, testing the model’s ability to generalize. Here, although HarMA retrieves two incorrect texts, each corresponding to a specific image. Interestingly, as shown on the far right, the retrieved image is also a forest. From a human evaluation perspective, the incorrectly retrieved results may provide more detailed descriptions than the correct text, while the original dataset annotation appears rather simplistic. The results retrieved by Full-ft CLIP, on the other hand, deviate significantly from the actual image semantics, with the primary semantics being squares (TOP-1), farmland (TOP-2), and industrial areas (TOP-4). We hypothesize that HarMA, by introducing downstream domain knowledge through adapters while retaining some of the original large-scale pre-training priors, is better equipped to handle the ``noise problem'' in small datasets compared to full fine-tuning.

\subsection{Text-to-Image Results Analysis}

\begin{figure}[htbp]
\vspace{-5pt}
\centering
\includegraphics[width=1.0\textwidth]{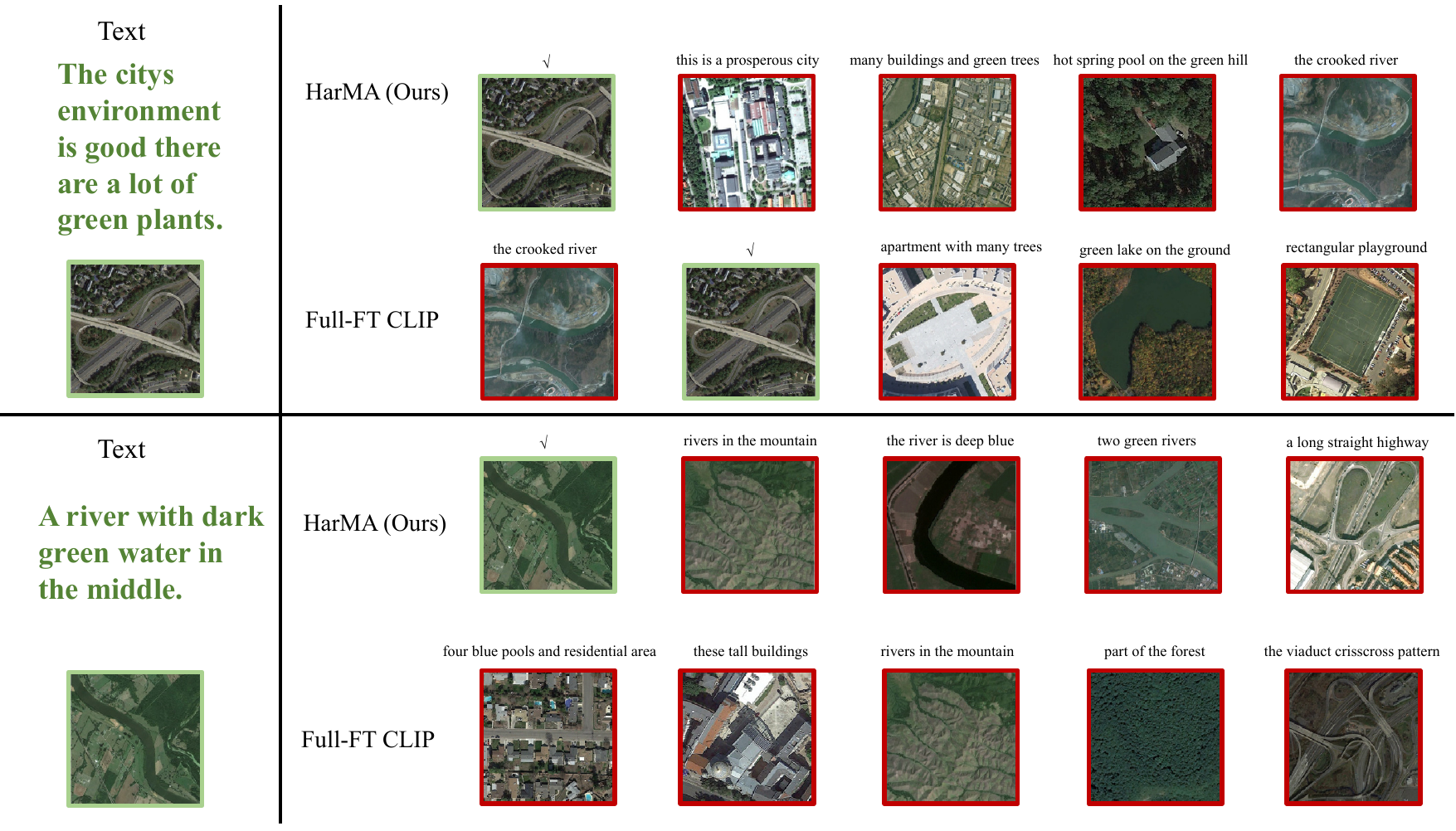}
\caption{\textbf{Text-to-Image Retrieval Visual Results.} On the left side, we show the query text, and the ground-truth image is displayed below. The top-5 retrieved images based on the query text are presented on the right side, where green boxes indicate matches and red boxes indicate mismatches. For the mismatched retrieved images, we identify the main semantics of their associated text above the images. It is worth noting that in the rsitmd dataset, the relationship between images and texts is one-to-many, meaning that Image-to-Text can retrieve multiple results, while Text-to-Image has only one correct result.}
\label{t2i res}
\end{figure}

Figure \ref{t2i res} illustrates the top-5 text-to-image retrieval results on the RSITMD test set for our HarMA and the fully fine-tuned CLIP models. Similar to the image-to-text case, our HarMA outperforms the fully fine-tuned CLIP overall.

Let us examine the first case. In the first row, the query image depicts ``a city with a lot of green plants.'' Our HarmA successfully retrieves a matching image in the top-1 result. Encouragingly, the top-2 to top-4 results are primarily semantically associated with ``city'' and ``green plants.'' Although the final top-5 result is semantically unrelated, the distorted river channel bears striking resemblance to the winding road in the query image. The second row shows the retrieval results from the fully fine-tuned CLIP, which, similar to the image-to-text case, focuses on irrelevant details (e.g., the green lake in top-4 and the green playground in top-5).

In the second case, both HarMA and the fully fine-tuned CLIP exhibit varying degrees of ``hallucination'' by associating the winding river with distorted roads. However, HarMA still outperforms the fully fine-tuned CLIP in terms of retrieval confidence and overall semantic relevance. For instance, HarMA still captures the overall semantics of ``A river with dark green water.''

In conclusion, the image-to-text and text-to-image retrieval results demonstrate HarMA's effectiveness in mitigating hallucinations and resisting noise. In the future, we aim to extend it to LLMs.

\subsection{Visual Results of Embedding Space}
In this subsection, we present the t-SNE \citep{van2008visualizing} visualizations of the image and text embeddings. To capture the highest-level semantics, we select the embeddings from the final transformer layer for t-SNE processing. We employ a CLIP model with a ViT-B-32 backbone and conduct experiments on the RSITMD test set.

\begin{figure}[htbp]
\centering
\includegraphics[width=1.0\textwidth]{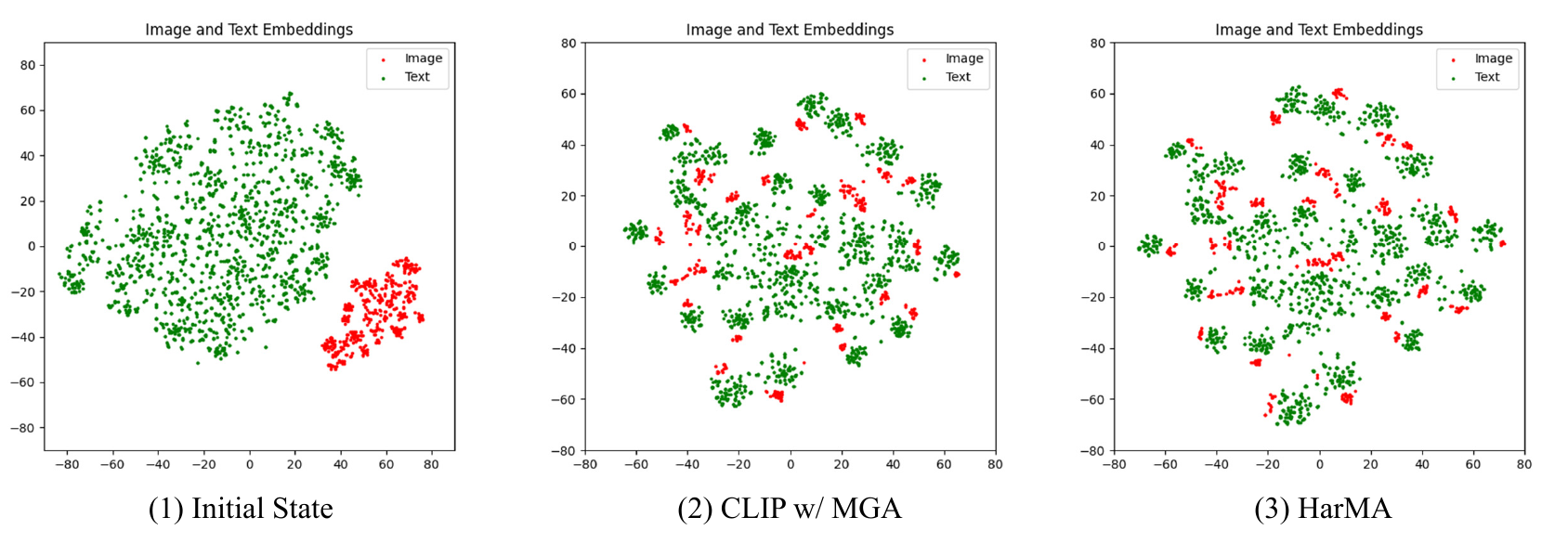}
\caption{The leftmost figure illustrates the visualization results of the CLIP model outputs without fine-tuning on downstream data. The central figure depicts the visualization of the CLIP outputs with MGA. The rightmost figure showcases the results of our HarMA framework.}
\label{tsne res}
\end{figure}

As shown in Figure \ref{tsne res}, the visualization of the CLIP model outputs without fine-tuning on downstream data reveals a significant distance between the embeddings of different modalities, indicating suboptimal modal alignment. The CLIP output results with MGA demonstrate excellent modal alignment; however, the distances within the same modality are excessively small (exhibiting partial clustering), potentially hindering the capture of fine-grained semantic differences. 

\begin{figure}[htbp]
\centering
\includegraphics[width=1.0\textwidth]{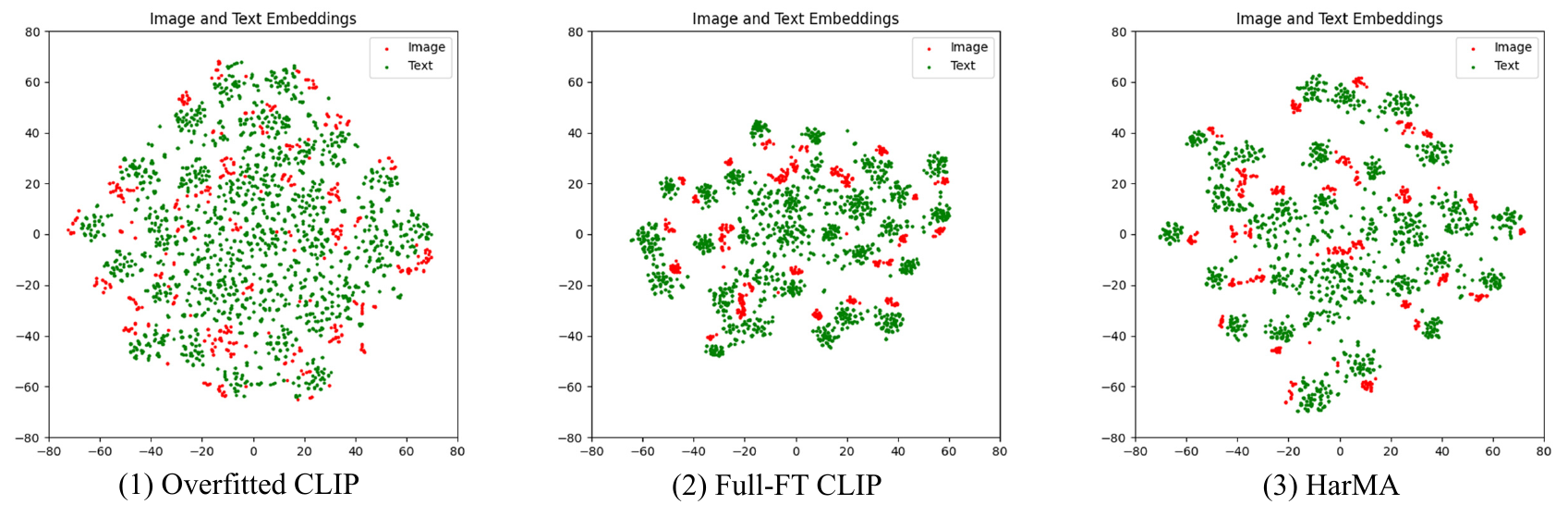}
\caption{\textbf{Visualization of the embedding space.} The first image illustrates the output of an overfitted CLIP model, the second image depicts the results from a fully fine-tuned CLIP model, and the third image presents the outputs after fine-tuning with the HarMA framework.}
\label{overfit}
\end{figure}

Figure \ref{overfit} presents the visualization results of the embedding space under various conditions. The retrieval performance is ranked as follows: HarMA $>$ Fully fine-tuned CLIP $>$ Overfitted CLIP. The overfitted CLIP model shows almost no alignment between visual and textual embeddings, resulting in the poorest retrieval performance. The fully fine-tuned CLIP model achieves basic modality alignment but still exhibits some clustering phenomena mentioned in Section \ref{intro}. Conversely, regardless of whether in Figure \ref{tsne res} or Figure \ref{overfit}, the output results of the complete HarMA framework not only exhibit robust modal alignment but also mitigate the aforementioned intra-modal clustering phenomenon, thereby validating the effectiveness of our proposed method.

\end{document}